\documentclass[runningheads]{llncs}

\usepackage[T1]{fontenc}

\usepackage{graphicx}
\usepackage{comment}
\usepackage{times}
\usepackage{epsfig}
\usepackage{graphicx}
\usepackage{wrapfig}
\usepackage{amsmath}
\usepackage{amssymb}
\usepackage{mathtools}
\usepackage{flexisym}
\usepackage{breqn}
\usepackage{nicefrac}
\usepackage{color}
\usepackage[dvipsnames]{xcolor}
\usepackage{booktabs}
\usepackage{multirow}
\usepackage{url}            
\usepackage{xspace}
\usepackage[inline]{enumitem} 
\usepackage{caption}
\usepackage{subcaption}
\usepackage{csquotes}
\usepackage{textcomp}
\usepackage{tipa}

\usepackage[activate={true,nocompatibility},final,tracking=true,kerning=true,spacing=true,factor=1100,stretch=10,shrink=10]{microtype}
\microtypecontext{spacing=nonfrench}

\usepackage{array} 
\newcolumntype{H}{>{\setbox0=\hbox\bgroup}c<{\egroup}@{}}

\usepackage[accsupp]{axessibility}  

\usepackage{hyperref}

\usepackage[capitalize]{cleveref}
\crefname{section}{Sec.}{Secs.}
\Crefname{section}{Section}{Sections}
\Crefname{table}{Table}{Tables}
\crefname{table}{Tab.}{Tabs.}

\newcommand{\mypara}[1]{\vspace{2pt}\noindent{\bf{#1}}}

\newcommand{\modelName}{ZS-A2T\xspace}

\DeclareMathOperator*{\argmax}{arg\,max}

\newif\ifreview
\reviewfalse

\ifreview
	\usepackage{lineno}

	\linenumbers
\fi

\hypersetup{
pdftitle={Zero-shot Translation of Attention Patterns in VQA Models to Natural Language},
pdfauthor={Leonard Salewski, A. Sophia Koepke, Hendrik P. A. Lensch, Zeynep Akata},
pdfkeywords={Zero-Shot Translation of Attention Patterns, Visual Question Answering (VQA), Large Language Models (LLM)}
}

\begin{document}


\def\SubNumber{24}

\def\GCPRTrack{Main Track}

\title{Zero-shot Translation of Attention Patterns in VQA Models to Natural Language}

\ifreview
\else

    \author{Leonard Salewski\inst{1}\orcidID{0000-0001-8531-3011} \and
    A.\ Sophia Koepke\inst{1}\orcidID{0000-0002-5807-0576} \and
    Hendrik P.\ A.\ Lensch\inst{1}\orcidID{0000-0003-3616-8668} \and
    \mbox{Zeynep Akata}\inst{1,2}\orcidID{0000-0002-1432-7747}}
    \authorrunning{L.\ Salewski et al.}
    %
    \institute{University of T{\"u}bingen, Tübingen AI Center \and
    MPI for Intelligent Systems\\
    \email{\{leonard.salewski, a-sophia.koepke, hendrik.lensch, zeynep.akata\}@uni-tuebingen.de}}
\fi

\maketitle              

\begin{abstract}
Converting a model's internals to text can yield human-understandable insights about the model.  
Inspired by the recent success of training-free approaches for image captioning, we propose \modelName, a zero-shot framework that translates the transformer attention of a given model into natural language without requiring any training. We consider this in the context of Visual Question Answering (VQA). 
\modelName builds on a pre-trained large language model (LLM), which receives a task prompt, question, and predicted answer, as inputs. The LLM is guided to select tokens which describe the regions in the input image that the VQA model attended to. 
Crucially, we determine this similarity by exploiting the text-image matching capabilities of the underlying VQA model.
Our framework does not require any training and allows the drop-in replacement of different guiding sources (e.g.\ attribution instead of attention maps), or language models.
We evaluate this novel task on textual explanation datasets for VQA, giving state-of-the-art performances for the zero-shot setting on GQA-REX and VQA-X.
Our code is available \href{https://github.com/ExplainableML/ZS-A2T}{here}.

\keywords{Zero-Shot Translation of Attention Patterns \and VQA.}
\end{abstract}

\section{Introduction}
Deep learning systems have become an integral part of our society, both in non-obvious applications, e.g.\ customer credit scoring, as well as in prominent applications, e.g.\ ChatGPT~\cite{Ouyang2022TrainingLM}.
Their impact on the lives of millions of people establishes the need to make these algorithms more transparent and accessible.
In the context of Visual Question Answering (VQA)~\cite{agrawal2015VQAVisualQuestion,goyal2017MakingVVQA}, methods for attribution~\cite{selvaraju2017grad} or attention visualization~\cite{abnar2020quantifying} aim to highlight the input image regions that are most relevant for the final decision.
However, the actual visual concepts that the model \enquote{saw} in the salient regions can remain obscure to the user.
In contrast, a natural language description of those visual concepts can be a more intuitive format for a human user.

There is a wide variety of approaches to determine image regions that were relevant for a model's output. 
Many of those methods do not require training and can be directly applied to a given model to generate attribution or attention visualizations.
As it is infeasible to train a dedicated model for translating method-specific visual explanations to natural language, we try to address the question: 
Can we convert the output of any attention or attribution method 
to natural language without any supervisory data?

\begin{wrapfigure}[15]{r}{.5\linewidth}
    \centering
    \vspace{-1.5em}
    \includegraphics[width=\linewidth]{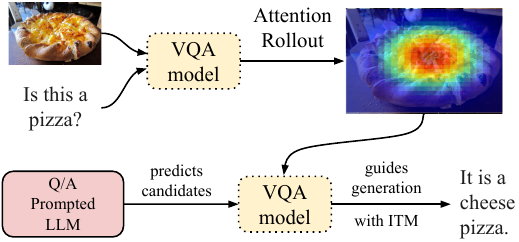}
    \caption{
    Our \modelName framework performs training-free translation of attention patterns from VQA models into natural language by combining a pre-trained large language model with language guiding with the VQA model's attention.
    }%
    \label{fig:teaser}
\end{wrapfigure}
Inspired by the impressive capabilities of pre-trained LLMs, we propose a \textit{zero-shot attention-to-text} (\modelName) framework which translates the internal attention of a transformer-based VQA model into natural language without requiring training (see \Cref{fig:teaser}). 
In particular, \modelName uses a LLM
that is steered by visual inputs corresponding to model attribution or attention visualizations.
We judge the visual relevance of the LLM's proposals with an image-text matching framework.
In contrast to related zero-shot image captioning methods like~\cite{Su2022LanguageMC,Tewel2021ZeroCapZI,Tewel2022ZeroShotVC}, \modelName does not exploit CLIP~\cite{Radford2021LearningTV}, whose image-text understanding would be different from the original task model.
Instead, we re-use the encoders of the underlying VQA model
for quantifying the agreement between the visual evidence and the candidate word.
This guides the text generation without introducing external input into the translation process.

We hypothesize that the content of the verbalizations of the attention patterns for VQA should capture the visual evidence that was used for VQA\@. This visual evidence should also be described in the corresponding textual explanations. Therefore, we evaluate the quality of our generated attention pattern translations on the VQA-X and GQA-REX datasets. 
Whilst naturally giving weaker results than methods that were trained to generate explanations in a fully supervised manner along with solving the VQA task, our proposed \modelName outperforms all related methods in this novel zero-shot attention pattern translation setting.
Additionally, the training-free setup of our method allows our approach to use different language models without any adaption or training.
Similarly, our framework works for any attention aggregation or visual attribution method that can be applied to the underlying VQA model.

To summarize, our contributions are:
1) We introduce the task of converting the internal structures of VQA models to natural language. 
2) Our proposed zero-shot \modelName framework is a simple, yet effective method for generating textual outputs with the guidance of the aggregated internal attention and text-image matching capabilities of the VQA model. 
3) \modelName can be utilized in conjunction with any pre-trained language model or visual attribution technique, and achieves state-of-the-art results on the GQA-REX and VQA-X textual explanation benchmarks in the zero-shot setting.

\vspace{-0.65ex}
\section{Related Work}
\mypara{Attention visualizations and visual attribution.}
Visual attribution methods commonly use backpropagation to trace a model's output to its input, e.g.\ by visualizing the (slightly modified) gradients on the corresponding input image~\cite{simonyan2013deep,springenberg2014striving,zeiler2014visualizing,smilkov2017smoothgrad,zhang2018top,bach2015pixel}. In addition to that, CAM~\cite{zhou2016learning}, Grad-CAM~\cite{selvaraju2017grad,selvaraju2019GradCAMVisual}, and variants thereof~\cite{draelos2020use,fu2020axiombased,Muhammad2020EigenCAMCA,Pillai2021ExplainableMW} use model activations in the attribution visualizations.
Different from visual attribution methods, perturbation-based methods slightly alter the input and record the resulting changes in the output~\cite{fong_iccv_2017,fong2019understanding}. Unfortunately, it is hard to quantify the quality of attribution visualizations~\cite{kim2022hive}. 
For modern transformer-based models, a number of studies~\cite{Jain2019AttentionIN,Wiegreffe2019AttentionIN,abnar2020quantifying,Chefer2021GenericAE} have investigated the extraction and visualization of attention scores.
In particular, attention rollout~\cite{abnar2020quantifying} determines the relevance of input tokens by following the flow of information that is captured in the attention scores and residual connections of the transformer network. However, it is not clear if the resulting visualizations are an intuitive way of explaining deep learning models to human users.
Our proposed \modelName framework offers a way of translating such outputs into natural language.

\mypara{Visual conditioning for text generation with LLMs.}
Several works have used LLMs for zero-shot image captioning~\cite{Tewel2021ZeroCapZI,Su2022LanguageMC,Tewel2022ZeroShotVC,Zeng2022SocraticMC,Wang2022ZeroshotIC}.
ZeroCap~\cite{Tewel2021ZeroCapZI} uses CLIP~\cite{Radford2021LearningTV} for updating the language model's hidden activations to match the image input.
Similarly, MAGIC~\cite{Su2022LanguageMC} combines the prediction of the language model, CLIPs rating and a degeneration penalty~\cite{Su2022ACFsimctg}. 
EPT~\cite{Tewel2022ZeroShotVC} only optimizes the hidden activations of a few selected pseudo-tokens for zero-shot video captioning. In contrast to these approaches, Socratic Models~\cite{Zeng2022SocraticMC} generates captions by conditioning the language model on CLIP-detected class names.
Another creative approach~\cite{Wang2022ZeroshotIC} finetunes a language model to process CLIP text embeddings which at test time can be replaced by CLIP image embeddings.
Similarly,~\cite{Li2023DeCapDC} finetunes with caption data and addresses the modality gap in a training-free manner.
Different from the two aforementioned methods~\cite{Wang2022ZeroshotIC,Li2023DeCapDC}, \modelName does not require any training (beyond the pre-trained models used) or external models for guiding the language generation, such as CLIP\@. In particular, the text generation in \modelName is controlled using the VQA model's internal attention along with exploiting the image-text matching capabilities of the same model.

\mypara{Textual explanations for VQA.}
Several VQA datasets with textual explanations have driven research on explaining outputs of models trained for VQA\@.
In particular, the VQA-X~\cite{hukpark2018MultimodalExplanationsJustifying} dataset extends a subset of VQA~v2~\cite{goyal2017MakingVVQA} with textual explanations obtained from humans.
In contrast, the VQA-E dataset~\cite{li2018vqa} automatically sources explanations from image captions.
Recently, the CLEVR-X~\cite{salewski2022clevr} and GQA-REX~\cite{Chen2022REXRA} extended the CLEVR~\cite{johnson2017CLEVRDiagnosticDataset} and GQA~\cite{hudson2019GQANewDataset} datasets with textual explanations by exploiting the corresponding scene graphs. In this work, we evaluate the verbalization of a model's internal attention on textual explanation datasets in the context of VQA, since textual explanations and visual attribution visualizations should both indicate the key features that influenced a VQA model's output.

\begin{figure*}[t]
    \centering
    \includegraphics[trim={0 .1em 0 0},clip,width=\linewidth]{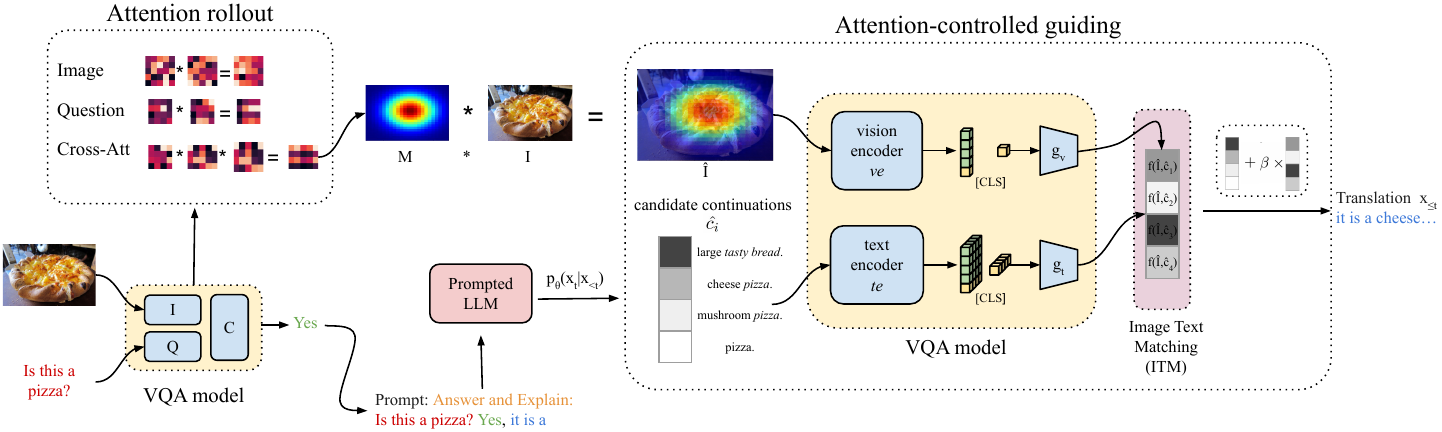}
    \caption{
        Our \modelName framework translates the attention rollout~\cite{abnar2020quantifying} of a VQA model into natural language.
        A pre-trained large language model is prompted with the {\color{Orange}task description}, the {\color{red}question $\mathcal{Q}$}, and the {\color{ForestGreen}predicted answer $a$}.
        The candidates for the {\color{NavyBlue}next token $x_t$} along with their continuations (\textit{italic}) are then re-ranked by re-using the image-text matching capabilities of the VQA model.
       We match the full sentences consisting of already {\color{NavyBlue}generated tokens}, the {\color{NavyBlue}candidates} and their respective {\color{NavyBlue}continuations} with the attention-masked image $\hat I$.
        This information is used to steer the output of a pre-trained language model without requiring any training.
    }%
    \label{fig:model}
\end{figure*}

\section{The \modelName Framework}\label{sec:framework}
In this section, we explain our proposed framework  which is visualized in \Cref{fig:model}. We propose the \modelName (\textbf{Z}ero-\textbf{S}hot \textbf{A}ttention \textbf{to} \textbf{T}ranslation) framework which converts the internal attention in a VQA model into natural language by translating the visualization of the aggregated attention in the transformer model into natural language without any supervision.
In particular, we prompt a pre-trained large language model for the translation task, and additionally provide the input question and the answer prediction from the VQA model.
The text generation is guided by the attention rollout result with the help of the VQA model itself.
Token by token, this approach converts visual explanations, i.e.\ visual attention patterns, into natural language.

\mypara{Pre-trained language model.}
Our \modelName framework exploits a pre-trained large language model as its text generator.
We condition the language model on two different inputs: (i) a task description, (ii) sample specific inputs (i.e.\ the question $\mathcal{Q}$, and the predicted answer $a$), and the already generated tokens.
We then generate the translation in an autoregressive setup by feeding each predicted token back to the language model.
It models the probability 
\begin{equation}\label{eq:lm-prediction}
p_{\theta}(x_t \mid \underbrace{x_0, \ldots, x_{l-1}}_{\text{task}}, \underbrace{x_{l}, \ldots, x_{l+s-1}}_{\text{sample ($\mathcal{Q}$, $a$)}},\underbrace{x_{l+s}, \ldots, x_{t-1}}_{\text{autoregressive}})    
\end{equation}
 for a token $x_t$ at step $t$, where $l$ is the number of task tokens and $s$ is the number of sample tokens.
 ${\theta}$ are the parameters of the pre-trained language model.

The probability of the next token $x_t$ defined in~\Cref{eq:lm-prediction} is independent of the input image.
As the pre-trained language model is conditioned on the question and answer, we expect it to predict tokens that fit the question and answer.
However, candidate tokens that are likely from a grammatical or statistical point of view, e.g.\ those that occur frequently in the training text corpus, will be ranked highly.
Thus, we argue that the language model predicts the correct tokens for the given question, image, and answer triplet, but not necessarily ranked correctly.
To accurately describe the visual concepts, we rely on the attention-controlled guiding of the VQA model.

\mypara{Attention-controlled guiding.}
To generate natural language that actually corresponds to the prediction of the VQA model, we condition the text generation on a masked version of the input image.
We obtain an attention-masked image by removing the image information that was not relevant for generating the VQA model's answer prediction $a$. The removal of information from the image is determined by the internal attention of the VQA model. For this, we use attention rollout which will be described in the following.

\mypara{Attention rollout.}
The attention of the given VQA model is aggregated based on attention rollout~\cite{abnar2020quantifying}. Here, we describe the rollout process for an encoder-decoder VQA model which interlinks the modalities with cross-attention.
However, different techniques for obtaining an attention-masked image could be used.

To determine the answer relevant image parts, we record the self-attention scores $A_{l}^{qq}$ (for the question tokens) and $A_{l}^{ii}$ (for the image tokens).
We also trace the cross-attention scores $A_{l}^{qi}$, which model the attention to the image tokens by the question tokens in layer $l$.
The attention scores are saved for each layer $l$ in the VQA model during inference.
As attention heads differ in importance~\cite{voita-etal-2019-analyzing} we only use the maximum activation $\bar{A}_l=\text{max}(A_l)$ across all heads.
Following~\cite{abnar2020quantifying,Chefer2021GenericAE}, we model how the attention flows throughout the network.
We successively multiply the attention scores of a layer with the rolled out values $R_l$ of the previous layer to obtain a final attention rollout map.
Additionally, we add the map of the previous layer $\tilde R_{l-i}$ to model the residual connection.
For self-attention, $R_0$ is initialized as a diagonal matrix since each token only contains its own information.
For cross-attention, $R_0^{qi}$ is initialized to zero, as the tokens have not been contextualized yet.

To roll out the self-attention for the question and visual tokens, we compute the following values for all layers which leverage self-attention before applying normalization,
\begin{equation}
     R_l^{qq} = R_{l-i}^{qq} + \bar{A}_l^{qq} \cdot R_{l-i}^{qq}, \qquad
    R_l^{ii} = R_{l-i}^{ii} + \bar{A}_l^{ii} \cdot R_{l-i}^{ii}.
\end{equation}
Additionally, we unroll the cross-attention to capture how the VQA model incorporates the visual information into the question representation:
\begin{equation}
    R_l^{qi} = R_{l-i}^{qi} + {R_{l-i}^{ii}}^\intercal \cdot \hat{A}_l \cdot R_{l-i}^{qq}.
\end{equation}
After modelling the residual connection, this captures how the attention rollout of both the question and image tokens is incorporated into the cross-attention.
For layers that mix self- and cross-attention we also account for the cross-modal information mixed in at the previous layer:
\begin{equation}
    R_l^{qi} = R_{l-i}^{qi} + \hat{A}_l^{qq} \cdot R_{l-i} ^{qi}.
\end{equation}
In the final layer, $R_L^{qi}$ models the importance of each input question token and each visual token for the VQA model's prediction.
We average $R_L^{qi}$ over all question tokens to obtain the final aggregated attention map $\bar{R}_L^{qi}$.
The threshold $\tau$ is applied to get a binary mask which is rescaled to the input image size to obtain $M$,
\begin{equation}
    M=\begin{cases}
    1, & \text{if $\tau < \bar{R}_L^{qi}$},\\
    0, & \text{otherwise}.
  \end{cases}
\end{equation}
The attention-masked image $\hat{I}$ is computed by eliminating irrelevant image parts $\hat{I}=M * I$, where $*$ is the element-wise multiplication.

\mypara{Visually guided text generation.}
In the following, we describe the attention-controlled visual guiding of the language decoding process in detail. 
The language generation at step $t$ starts with the pre-trained language model predicting the probability distribution over all tokens in the vocabulary.
We only consider the $\text{top-}k$ predictions $c_1,\ldots,c_k$ for the token $x_t$ and refer to those as \emph{candidates}.
This subset can be selected for two reasons: a) as argued above it is plausible that a sufficiently large language model conditioned on the question and answer ranks plausible words highly, and b) the weighted sum described in \Cref{eq:next-token} does not change for sufficiently large $k$ as $p_\theta$ is a result of a softmax operation, yielding small values for non-top activations.

In contrast to previous works~\cite{Tewel2021ZeroCapZI,Tewel2022ZeroShotVC,Su2022LanguageMC} which score incomplete sentences with an external image-text matching model (e.g.\ CLIP), we score a completed sentence with the same task model whose attention patterns we aim to convert to text.
Our approach relies on the fact that image-text matching losses used by VQA models are commonly supervised with complete sentences instead of sentence fractions.

We let the language model complete the sentence for each candidate $c_i$ separately based only on the predictions of the language model conditioned on the already generated tokens $x_{\leq t}$ and the candidate $c_i$.
Each continuation is terminated at the first \enquote{.}, such that the previously generated tokens, the candidate, and the continuation form a complete sentence $\hat{c_i}$.
We do not include the question or answer as parts of the candidates.
We use top-$p$ sampling to filter the predictions of the language model to get a plausible continuation. 
Next, the completed sentences $\hat{c_i}$ are ranked by performing image-text matching.

The attention controlled guiding is executed by feeding the text input $\hat{c_i}$ and the masked image $\hat{I}$ as inputs to the VQA model's uni-modal encoders. In particular, we proceed with the contextualized embedding of the \texttt{[CLS]} token $ve(\hat{I})$ of the vision encoder and the corresponding \texttt{[CLS]} token $te(\hat{c_i})$ of the text encoder. Both tokens are projected into a joint space using the linear maps $g_v$ and $g_t$ provided by the VQA model. This allows to measure the quality of the matching of the image-text pairs.

The matching quality $f(\hat{I},\hat{c}_i)$ of a candidate sentence $\hat{c}_i$ with respect to the attention rollout-masked image $\hat{I}$ is determined by first computing the cosine similarity $\text{CosSim}(\cdot,\cdot)$ of all possible image-sentence matches,
\begin{equation}
\begin{split}
    f(\hat{I},\hat{c}_i)
    =
    \frac{
        e^{\kappa \cdot \text{CosSim} \left(g_v(ve(\hat{I})), g_t(te(\hat{c}_i)) \right)}
    }{
        \sum_{j \in 1,\ldots,k}e^{\kappa \cdot \text{CosSim} \left(g_v(ve(\hat{I})), g_t(te(\hat{c}_j))\right)}
    }.
\end{split}
\end{equation}
Then, we apply a temperature $\kappa$ in the softmax operation which affects the sharpness of the distribution. 
Very small values of $\kappa$ would approximate a uniform distribution, largely disabling the influence of the visual guiding component.
On the other hand, too large values may overemphasize highly salient visual concepts, at the cost of grammatical issues (due to overruling the language model).
Thus, $\kappa$ is manually chosen, such that the distribution of the predictions $f$ roughly matches the language model $p_{\theta}$. 

To determine the next token, we compute a weighted sum of the prediction of the language model $p_\theta$ and the matching quality $f$ obtained with the VQA model.
Thus, at time step $t$, the next token $x_t$ is computed according to:
\begin{equation}\label{eq:next-token}
\begin{split}
    x_t
    =
    \argmax_{i \in 1, \ldots, k}
    \Big\{
        p_{\theta} \left(c_i \mid x_0, \ldots, x_{<t} \right) +  \beta \cdot  f_{} \left(\hat{I}, \hat{c_i} \right)
    \Big\},
\end{split}
\end{equation}
where $\beta$ is a scalar weighting factor.
After selecting the next token $x_t$, we append it to the original language model prompt and repeat the above process for generating tokens at step $t+1$ until reaching a stopping criterion (either an \texttt{[EOS]} token, or a period).

\section{Experiments}
In this section, we describe our experimental setup, including the pre-trained models used in \modelName, the datasets used, and the evaluation metrics. Experimental results generated by \modelName for the zero-shot textual explanation task in the context of VQA are compared to four related training-free methods on the VQA-X and GQA-REX datasets. Although we do not claim that our generated texts are explanations, we do believe that these datasets are well suited for evaluation. 
Finally, we investigate individual components in our framework in detail, such as the attention-controlled guiding, and the language model prompting before providing qualitative results for \modelName.

{
\renewcommand{\arraystretch}{1.2}
\begin{table*}[t]
    \centering
    \resizebox{.95\linewidth}{!}{
        \begin{tabular}{c | l H H H c c c c c  c  H H H c c c c c}
            \toprule
            \multicolumn{2}{c}{} & \multicolumn{8}{c}{\textbf{GQA-REX}~\cite{hudson2019GQANewDataset,Chen2022REXRA}}  & & \multicolumn{8}{c}{\textbf{VQA-X}~\cite{hukpark2018MultimodalExplanationsJustifying}}  \\
            \cmidrule(lr){3-10} \cmidrule(lr){12-19}
           Setting & \textbf{Framework}$\downarrow$ &  B1 & B2 & B3 & B4 & M & RL & C & S  & &  B1 & B2 & B3 & B4 & M & RL & C & S \\
           \midrule
            \multirow{6}*{Zero-shot}& ZeroCap \textsubscript{GPT-2}~\cite{Tewel2021ZeroCapZI}$^*$ &  10.6 & 4.7 & 2.4 & 1.4 & 4.6 & 12.3 & 16.9 & 5.3 && 10.3 & 4.3 & 1.7 & 0.7 & 4.7 & 14.0 & 5.8 & 2.0  \\
            & EPT \textsubscript{GPT-2}~\cite{Tewel2022ZeroShotVC}$^*$ & 4.5 & 1.0 & 0.3 & 0.0 & 3.3 & 3.2 & 2.6 & 2.8 && 20.8 & 6.4 & 2.3 & 0.9 & 6.5 & 14.9 & 6.7 & 2.9 \\
            & MAGIC \textsubscript{GPT-2}~\cite{Su2022LanguageMC}$^{*}$ & 18.5 & 7.9 & 4.0 & 2.3 & 10.8 & 18.8 & 41.1 & 18.8 && 26.1 & 9.0 & 3.1 & 1.0 & 8.8 & 19.3 & 10.6 & 7.1  \\
            & MAGIC \textsubscript{OPT 6.7B}~\cite{Su2022LanguageMC}$^{*}$ & 23.0 & 10.8 & 5.6 & 3.3 & 11.6 & 22.2 & 48.8 & 21.4 && 27.5 & 10.6 & 4.3 & 1.9 & 9.5 & 20.5 & 14.7 & 8.9\\
            & Socratic Models \textsubscript{OPT 6.7B}~\cite{Zeng2022SocraticMC}$^{*}$ & 20.3 & 9.7 & 5.4 & 3.3 & 14.1 & 22.8 & 40.5 & 19.3 && 31.3 & 15.4 & 7.6 & 3.6 & 12.8 & 25.7 & 19.9 & 10.1\\
            \cmidrule(lr){2-19}
            & \modelName \textsubscript{OPT 6.7B} (ours)  & \bf 38.5 & \bf  22.6 & \bf 14.7 & \bf 10.2 & \bf 18.2 & \bf 35.0 & \bf 113.5 & \bf 31.4 && \bf 41.6 & \bf 25.5 & \bf 14.7 & \bf 8.5 & \bf 13.8 & \bf 34.2 & \bf 38.1 & \bf 10.5\\
            \midrule
            \midrule
            \multirow{2}*{Supervised} & NLX-GPT~\cite{Sammani2022NLXGPTAM} \textsubscript{GPT-2} & - & - & - & - & - & - & - & - & & - & - & - & 23.8 & 20.3 &  47.2  & 89.2 & 18.3 \\ 
            & VisualBert-REX~\cite{Chen2022REXRA} \textsubscript{LSTM}   & - & - & - & 54.6 & 39.2 & 78.6 & 464.2 & 46.8 &  & - & - & - & - & - & - & - & -\\
            \bottomrule
        \end{tabular}
    }
    \caption{Zero-shot model attention to natural language translation results in the context of VQA evaluated on GQA-REX and VQA-X. We report Bleu-4~(B4), Meteor~(M), Rouge-L~(RL), Cider~(C), and Spice~(S). Higher is better for all reported metrics. 
    State-of-the-art performances with textual explanation generation methods in the context of VQA are included for reference (supervised). 
    $^*$We adapted the related frameworks to the attention to natural language translation task, and also provided those methods with privileged access to the VQA ground-truth answers.}%
    \label{tab:sota-results}
\end{table*}
}

\mypara{Experimental setup.}\label{suse:experimental_setup}
We used the OPT~\cite{Zhang2022OPTOP} pre-trained language models for language generation.
The transformer-based ALBEF~\cite{Li2021AlignBF} finetuned for VQA, served as our underlying VQA model. The weights for the projection layers are loaded from the non-finetuned, pre-trained ALBEF model, which was trained on an image-text matching objective. Both ALBEF variants were \emph{not} trained to generate texts from the datasets that we test on (VQA-X and GQA-REX).
The threshold for the attention rollout scores was set to $\tau=\nicefrac{200}{256}$. 
For guiding, we determine the matching quality for the top-$k$ candidate tokens as $k=45$ and our continuations are sampled with $p=0.15$.
The guiding temperature is set to $\kappa=100$.
We set the scalar weighting factor between the language model and the attention-controlled guiding to $\beta=0.7$.
To maintain a clean zero-shot protocol, we selected the hyperparameters on the validation split of VQA-X, which is disjoint from the subsets used for the final evaluation.
We applied the same settings for GQA-REX without any further tuning.

\mypara{Datasets.}
To evaluate all methods, we used the VQA-X~\cite{hukpark2018MultimodalExplanationsJustifying} and GQA-REX~\cite{hudson2019GQANewDataset,Chen2022REXRA} datasets for textual explanations in the context of VQA\@. 
VQA-X\footnote{Licensed under the BSD-2 license.} extends a subset of non-trivial VQAv2~\cite{goyal2017MakingVVQA} questions with human-generated natural language explanations for the ground-truth answer.
GQA-REX\footnote{Licensed under the MIT license.} contains explanations for a subset of the real-world visual reasoning question answering task posed in the GQA dataset. 
GQA-REX contains one explanation per question-image pair.
As two of the related approaches have slow inference speeds ($\approx25$s and $\approx70$s per sample~\cite{Tewel2021ZeroCapZI,Tewel2022ZeroShotVC} on an NVIDIA v100 GPU), we evaluate all methods on a subset of the test set containing 2000 samples. This is comparable in size to VQA-X's test set.
Our approach only needs 7.2s per sample, and we compare inference speeds of all methods in Section C of the supplementary material.

\mypara{Evaluation metrics.}
Unless stated otherwise, all models are evaluated in the zero-shot setting, i.e.\ without any training, on the test sets of the respective datasets using commonly reported natural language generation metrics similar to~\cite{hukpark2018MultimodalExplanationsJustifying}.
In particular, we report Bleu-4~(B4)~\cite{papineni2001BLEUMethodAutomatic}, Meteor~(M)~\cite{banerjee2005METEORAutomaticMetric}, Rouge-L~(RL)~\cite{lin2004ROUGEPackageAutomatic}, Cider~(C)~\cite{vedantam2015CIDErConsensusbased}, and Spice~(S)~\cite{anderson2016spice} scores. 
These metrics aim to capture the semantic overlap between sentences by measuring (modified) precision~\cite{papineni2001BLEUMethodAutomatic} and recall~\cite{papineni2001BLEUMethodAutomatic,banerjee2005METEORAutomaticMetric,lin2004ROUGEPackageAutomatic} of $n$-grams.
Additionally, generalizations of $n$-grams like stemming~\cite{banerjee2005METEORAutomaticMetric,vedantam2015CIDErConsensusbased}, measures of sentence fragmentation~\cite{banerjee2005METEORAutomaticMetric} or tf-idf weighting~\cite{vedantam2015CIDErConsensusbased} are applied to better match human judgement of sentence similarity.

\subsection{Comparing to Related Frameworks}\label{suse:experimental_results}
To evaluate the quality of the generated translations, we show experimental results on VQA-X and GQA-REX\@.
As there are, to the best of our knowledge, no related works that translate VQA attention patterns into natural language, we adapted a number of zero-shot image captioning methods.
For fair comparison, we modified all related works by prompting them with the question and ground-truth answer, as they do not have a dedicated VQA module. This favors the related works, as our framework may base its translation on a wrongly predicted answer.
Additionally, we show other common evaluation schemes in Section B of the supplementary material.

First, we compare to the zero-shot image captioning works~\cite{Tewel2021ZeroCapZI,Su2022LanguageMC,Tewel2022ZeroShotVC,Zeng2022SocraticMC}.
For a fair comparison, we adapted the two stronger models (MAGIC and Socratic Models) to operate with the same language model as the one used in our framework (OPT 6.7B).
\modelName outperforms all four related approaches for zero-shot translation of attention patterns into natural language by wide margins (see \Cref{tab:sota-results}).
Interestingly, the optimization-based approaches ZeroCap and EPT show relatively weak performances (on VQA-X more so than on GQA-REX). 
This suggests that longer, more complicated prompts (in contrast to the ones used for image captioning) make it hard to optimize helpful starting parameters for the next token prediction (ZeroCap) or next sentence prediction (EPT).

MAGIC gives stronger results, but it is still largely outperformed by our \modelName framework.
Similarly, Socratic Models (SMs), the strongest of the adapted related works, is outperformed in all metrics.
The generated sentences by \modelName exhibit greater word-by-word overlap with the ground-truth references than those of the related approaches. In particular, this is indicated by the $n$-gram-based metric Bleu-4 for which \modelName obtains a score of 8.5 compared to 3.6 for SMs on VQA-X.
Similarly, \modelName is stronger than SMs in terms of Meteor with 18.2 compared to 14.1 on GQA-REX and 13.8 compared to 12.8 on VQA-X.
The same pattern holds true for all other metrics.

For context, we also list results with the recently published supervised models NLX-GPT\cite{Sammani2022NLXGPTAM} for VQA-X and VisualBert-REX~\cite{Chen2022REXRA} for GQA-REX\@.
They both employ joint multi-modal transformer models to predict the sentences of the respective datasets.
Whilst not strictly comparable, since we translate only the question relevant image regions, it is still interesting to note that our 5-shot variant (c.f.\ \Cref{tab:in-context-learning}) significantly shrinks the gap to the supervised models which used 31k VQA-X and 128k GQA-REX training samples respectively, whereas our model does not require any training or just a few in-context examples.

\subsection{Ablation Studies on Guiding Inputs}
In this section, we study the impact of using different input images in our attention-controlled guiding (see \Cref{sec:framework}) as well as the effect of guiding with completed sentences (in contrast to incomplete sentences).
In addition to this, we provide results for using different visual explanation methods in \modelName, i.e.\ for visual attribution and perturbation methods.
We show additional ablations for the attention thresholding parameter $\theta$ and the guiding temperature $\kappa$ in Section D of the supplementary material.

\begin{wraptable}[10]{r}{.5\linewidth}

\vspace{-2.25em}
    \centering
    \resizebox{\linewidth}{!}{
        \begin{tabular}{l H H H c c c c c c }
            \toprule
            Guiding Input & B1 & B2 & B3 & B4 & M & RL & C & S \\
            \midrule
            Full Image & 41.5 & 25.2 & 14.3 & 8.1 & 13.7 & 34.1 & 37.6 & \bf 10.8\\
            No Continuation & 37.0 & 21.1 & 11.1 & 6.2 & 12.5 & 31.1 & 28.2 & 9.4 \\
            \midrule
            \modelName (Rel.\ Masking + Cont.) & \bf 41.6 & \bf 25.5 & \bf 14.7 & \bf 8.5 & \bf 13.8 & \bf 34.2 & \bf 38.1 & 10.5\\
            \bottomrule
        \end{tabular}
   }
    \caption{
        Ablating the guiding input and text continuation on VQA-X~\cite{hukpark2018MultimodalExplanationsJustifying}. 
        Our \modelName model uses an attention-masked image, obtained from attention rollout~\cite{abnar2020quantifying}, and text continuation.
    }%
    \label{tab:image-masking-ablation}
\end{wraptable}
\mypara{Influence of attention masking.}
\Cref{tab:image-masking-ablation} shows the effect of using an attention-masked image in the visual guiding.
Restricting the guiding of the language generation to the attention-masked image improves the language generation in terms of the Bleu-4, Meteor, Rouge-L, and Cider metrics. Interestingly, the Spice metric is slightly higher (10.8 vs.\ 10.5) when using the full image for guiding.

\mypara{Influence of using text continuations.}
In \Cref{tab:image-masking-ablation}, we also investigate the effect of using the language model to generate text continuations, so that the guiding component can judge completed sentences.
Using the continuations increases all metrics, e.g.\ Cider from 28.2 to 38.1.
We hypothesize that this happens for two reasons.
First, it reduces the distribution shift between the contrastive image-text matching training of the VQA model.
Second, it allows the guiding to judge whether a greedy selection of the visually grounded token at step $t$ may lead to a completed sentence that is not visually supported.

\begin{table}[t]
    \begin{subtable}{.53\linewidth}
      \centering
        \resizebox{\linewidth}{!}{
        \begin{tabular}{l H H H c c c c c }
            \toprule
            Attribution Method & B1 & B2 & B3 & B4 & M & RL & C & S  \\
            \midrule
            Att.\ GradCAM~\cite{selvaraju2017grad,Li2021AlignBF} & 40.4 & 23.9 & 13.3 & 7.5 & 13.3 & 32.8 & 33.0 & 9.2 \\
            EigenGradCAM~\cite{Muhammad2020EigenCAMCA} & 40.0 & 23.7 & 13.2 & 7.5 & 13.1 & 32.8 & 32.3 & 9.2 \\
            XGradCAM~\cite{fu2020axiombased} & 39.2 & 22.6 & 12.2 & 6.7 & 12.4 & 32.1 & 30.1 & 9.1 \\
            GradCAMElementwise~\cite{Pillai2021ExplainableMW} & 40.1 & 23.5 & 12.9 & 7.2 & 12.8 & 32.5 & 30.7 & 9.0 \\
            HiResCAM~\cite{draelos2020use} & 39.4 & 23.0 & 12.7 & 7.0 & 12.7 & 32.5 & 31.0 & 9.0 \\
            \midrule
            RISE~\cite{petsiuk2018rise} & 40.7 & 24.3 & 13.5 & 7.5 & 13.3 & 33.1 & 33.5 & 9.9\\
            \midrule
            \modelName (attention rollout~\cite{abnar2020quantifying}) & \bf 41.6 & \bf 25.5 & \bf 14.7 & \bf 8.5 & \bf 13.8 & \bf 34.2 & \bf 38.1 & \bf 10.5\\
            \bottomrule
        \end{tabular}
   }
    \end{subtable}%
    \hfill
    \begin{subtable}{.445\linewidth}
      \centering
        \resizebox{\linewidth}{!}{
        \begin{tabular}{l H H H c c c c c c }
            \toprule
            LM (\#Params)  & B1 & B2 & B3 & B4 & M & RL & C & S  \\
            \midrule
            GPT-2 (125M) & 33.5 & 17.0 & 8.0 & 3.6 & 11.0 & 26.6 & 19.8 & 7.7  \\
            \midrule
            OPT (125M) & 31.5 & 16.4 &  7.5 & 3.4 & 10.7 & 26.5 & 18.5 &  7.1\\
            OPT (350M) & 35.5 & 18.8 &  9.1 & 3.9 & 11.6 & 27.9 & 20.2 &  7.9\\
            OPT (1.3B) & 39.7 & 26.6 & 13.0 & 7.1 & 13.0 & 32.8 & 28.9 &  9.2\\
            OPT (2.7B) & 40.4 & 23.7 & 12.8 & 7.1 & 13.3 & 32.5 & 31.0 & 10.1\\
            OPT (6.7B) & \bf 41.6 & \bf 25.5 & \bf 14.7 & \bf 8.5 & \bf 13.8 & \bf 34.2 & \bf 38.1 & \bf 10.5\\
            \bottomrule
        \end{tabular}
   }
    \end{subtable} 
\caption{Ablating different attribution and perturbation methods (left) and type and size of the pre-trained language models (right) in our \modelName framework on VQA-X~\cite{hukpark2018MultimodalExplanationsJustifying}.}%
\label{tab:attribution-and-lm-size}
\end{table}

\mypara{Different visual explanation methods.}
In addition to using attention rollout to determine relevant image parts (described in \Cref{sec:framework}), we use five other attribution methods and the perturbation-based visual explanation method RISE~\cite{petsiuk2018rise}.
\Cref{tab:attribution-and-lm-size} (left) showcases that our approach can handle conceptually different visual explanation methods.

The backpropagation-based techniques from the GradCAM family build on the attention probabilities and their respective gradients~\cite{selvaraju2017grad,Li2021AlignBF} from the same layer of the VQA model ($L=11$).
Thus, the generated sentences using the different methods EigenGradCAM~\cite{Muhammad2020EigenCAMCA}, XGradCAM~\cite{fu2020axiombased}, GradCAMElementwise~\cite{Pillai2021ExplainableMW} and HiResCAM~\cite{draelos2020use} are of very similar quality in terms of the NLG metrics.
Overall, their scores are lower than the scores we obtain when using attention rollout~\cite{abnar2020quantifying}

We additionally show results with the input perturbation-based method RISE~\cite{petsiuk2018rise}.
It filters the images applied to the given VQA model and evaluates the VQA model multiple times whilst randomly masking parts of the image.
The final importance map is obtained by summing the random masks weighted by the predicted class probability.
This indicates the parts in the input that are salient for the VQA models' prediction.
We find that the texts generated for RISE are slightly worse than those obtained with attention rollout based attribution method (e.g.\ Bleu-4 7.5 vs.\ 8.5 (ours)).

Overall, the visual explanation methods' different abilities to identify the correct relevant image regions is reflected in the quality of the translations to natural language. 
Attention rollout~\cite{Chefer2021GenericAE} generally outperforms GradCAM~\cite{selvaraju2019GradCAMVisual} in identifying relevant image regions and the same pattern is found in the quality of the translations.
Thus, we conclude that our framework is not tied to a specific visual explanation method, and allows the drop-in replacement of different visual explanation methods for guiding the language generation.

\subsection{Language Models}
Our framework can be used with  different language models without any changes in the setup (c.f.\ \Cref{tab:attribution-and-lm-size} (right)).
Here, we analyze the performance of \modelName for different language models.
We demonstrate that our framework even outperforms other related approaches when using a pre-trained GPT-2 language model. For example, it achieves a Rouge-L value of 26.6 on VQA-X, whereas the previous best model with a GPT-2 backbone (MAGIC) only achieves 19.3 with the same language model.
Furthermore, we clearly outperform MAGIC on the $n$-gram metric Bleu-4 (3.6 vs.\ 1.0), as well as on Meteor, Rouge-L, Cider, and SPICE (see \Cref{tab:sota-results}).
This can be attributed to our temperature $\kappa$, which allows for better balancing of the two terms in \Cref{eq:next-token}.

Next, we analyze the impact of the size of the pre-trained language models. Our \modelName framework does benefit from larger, more powerful language models.
The performance increase with larger size is consistent across all metrics, e.g.\ Spice goes from 7.1 to 9.2 when using the OPT model with 125M vs.\ 1.3B parameters.
Using the 6.7B model boosts the Spice performance to 10.5. 
This suggests that high-quality candidate proposals are beneficial for the generated sentences.
Additionally, it also showcases a benefit of our training-free approach: Large or newer language models can be swapped in without additional cost.

\subsection{Prompt Ablations}
Here, we analyze the effect of different input prompts for the pre-trained language model on the generated text outputs (c.f.\ \Cref{tab:prompt-ablation-results}).

\begin{wraptable}[14]{r}{.525\linewidth}
\vspace{-2em}
    \centering
    \resizebox{\linewidth}{!}{
        \begin{tabular}{l H H H c c c c c }
            \toprule
            Prompt & B1 & B2 & B3 & B4 & M & RL & C & S  \\
            \midrule
            \textlangle{}q\textrangle{}? the answer is \textlangle{}a\textrangle{} because & 30.9 & 15.6 & 7.5 & 3.7 & 10.3 & 25.7 & 21.6 & 8.4 \\
            Q: \textlangle{}q\textrangle{}? A: \textlangle{}a\textrangle{}. E: & 31.0 & 15.1 & 7.4 & 3.5 & 11.0 & 23.2 & 22.6 & 9.6 \\
            Q: \textlangle{}q\textrangle{}\texttt{\textbackslash n}? A: \textlangle{}x\textrangle{}\texttt{\textbackslash n}. E:  & 32.4 & 16.2 & 8.4 & 4.2 & 11.2 & 23.7 & 23.0 & 9.6 \\
            Explain the A: \textlangle{}q\textrangle{}? The A is \textlangle{}a\textrangle{} because & 39.0 & 22.8 & 12.4 & 6.9 & 13.0 & 32.3 & 31.7 & 9.9 \\
            A and Explain: \textlangle{}q\textrangle{}?\texttt{\textbackslash n} The A is \textlangle{}a\textrangle{} because & 40.5 & 23.9 & 13.5 & 7.6 & 13.5 & 32.9 & 35.8 & \bf 10.6\\
            \midrule      
            A and Explain: \textlangle{}q\textrangle{}? The A is \textlangle{}a\textrangle{} because  & \bf 41.6 & \bf 25.5 & \bf 14.7  & \bf 8.5 & \bf 13.8 & \bf 34.2 & \bf 38.1 & 10.5\\
            \bottomrule
        \end{tabular}
    }
    \caption{
        Ablating different input prompts on VQA-X~\cite{hukpark2018MultimodalExplanationsJustifying}. \textlangle{}q\textrangle{} and \textlangle{}a\textrangle{} are placeholders for the question and answer. Q denotes \enquote{Question}, A \enquote{Answer}, and E \enquote{Explanation}.
        \texttt{\textbackslash n} denotes a new line.
        We show the full prompt templates in Section A of the supplementary material.
    }%
    \label{tab:prompt-ablation-results}
\end{wraptable}
Unsurprisingly, providing no task description (row 1) 
in the input prompt gives the worst outputs.
Removing the task description (\enquote{Answer and explain:}) yields significantly worse results with Bleu-4 decreasing from $8.5$ to $3.7$, suggesting that the task description is crucial for performance.

Inputting the task in a more structured way (\enquote{Question: \textlangle{}q\textrangle{}? Answer: \textlangle{}a\textrangle{}. Explanation:}) increases the language quality only slightly compared to using no task description.
However, the same structured prompt extended by newline characters \texttt{\textbackslash n} shows increased or similar metric values for all metrics.
A further improvement is achieved by using a meaningful task description (\enquote{Explain the answer:}) in the input prompt.
We used the best prompt (\enquote{Answer and explain:}) in \modelName.

\mypara{Impact of in-context generation ($n$-shot prompting).}
We analyze the effect of prefixing the context with complete examples of questions, answers, and their respective explanations. This enables the language model to better understand the task, as it can see some examples before generating text~\cite{brown2020language}.

\renewcommand{\arraystretch}{1.4}
\begin{wraptable}[9]{r}{.4\linewidth}
    \centering
    \vspace{-3em}
   \resizebox{\linewidth}{!}{
        \begin{tabular}{l ccc H H H c c c c c c }
            \toprule
            Model && $n$ && B1 & B2 & B3 & B4 & M & RL & C & S  \\
            \midrule
           \modelName && 0  &&  41.6 & 25.5 & 14.7 &  8.5 &  13.8 &  34.2 &  38.1 &  10.5\\
           \modelName  && 1  && 44.0 & 26.7 & 16.1 & 9.8 & 14.5 & 34.8 & 42.7 & 11.7 \\
           \modelName  && 5 &&  46.8 &  30.4 &  19.1 &  11.9 &  15.3 &  37.5 & 49.6 &  12.4\\
            \bottomrule
        \end{tabular}
  }
    \caption{
        In-context learning with $n$ examples on the VQA-X dataset.}%
    \label{tab:in-context-learning}
\end{wraptable}
We experiment with up to $n=1$ and $n=5$ randomly sampled examples from the training set of the respective dataset which are prepended to the context (see \Cref{tab:in-context-learning}).
The full prompts for this setup are included in Section E of the supplementary material.

Using a single example improves the language quality already. For $n=1$, the natural language generation scores are on average 7.2\% higher than for $n=0$.
By just prefixing five in-context learning examples, the generation quality increases on average by 19.3\% over providing no examples.
Qualitatively, we find that the language model does not merely copy or modify the texts of the examples when the question and/or answer match, but instead 
allows to adjust the candidate predictions for the generated sentences accordingly.
Additionally, the generated texts better match the language biases in the datasets (e.g.\ on VQA-X many samples start with \enquote{the} or \enquote{there}).

\subsection{Qualitative Results}
\begin{figure*}[t]
    \centering
    \includegraphics[width=\linewidth]{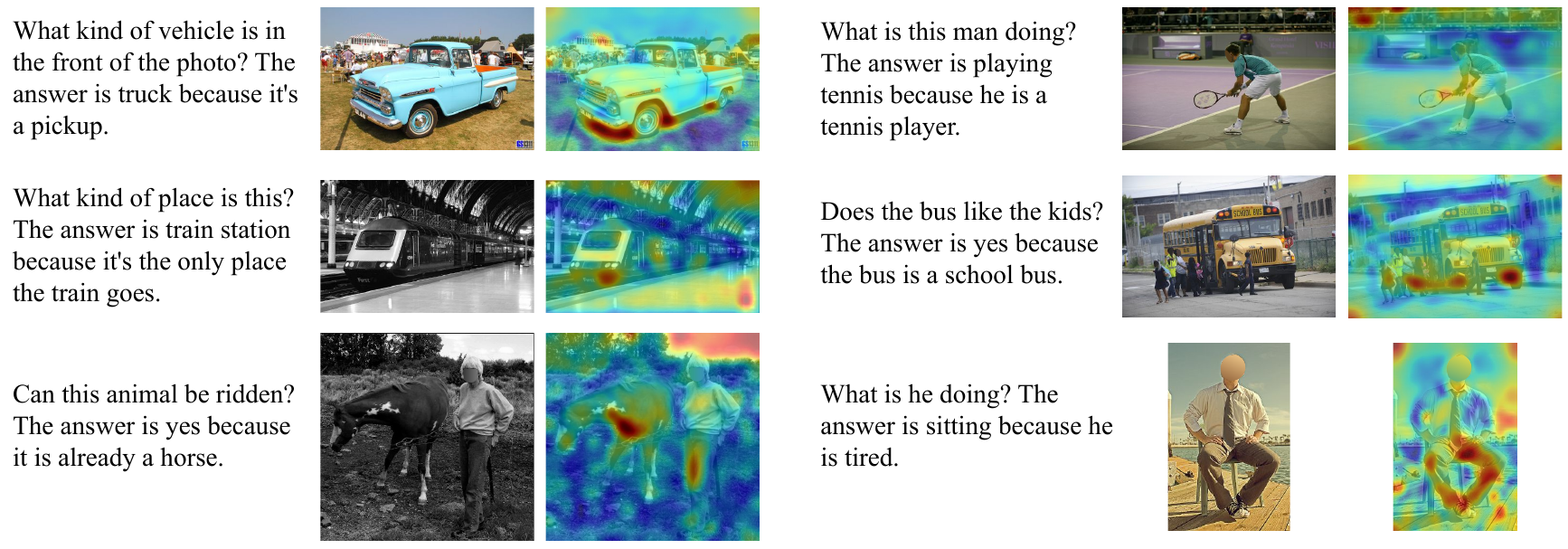}
    \caption{
        Qualitative results for \modelName on VQA-X, showing the input image, question, attention rollout~\cite{abnar2020quantifying} output, and generated text. Red indicates higher relevance. The generated text indeed mentions the visual concepts detected by the VQA model.
        The face regions have been deliberately occluded in this figure.
    }%
    \label{fig:qualitative-examples}
\end{figure*}
We provide qualitative examples for natural language translations generated with \modelName for VQA-X in \Cref{fig:qualitative-examples}.
The attention rollout map is superimposed onto the original image.
For each of the examples we show the question, predicted answer and generated sentence that translates the attention patterns into natural language. 
We can observe that attention rollout, used for the attention-controlled guiding, selects relevant image regions that plausibly correspond to the input question.
In the bottom left example the main attention is directed towards the \emph{horse}.
The generated sentences are fluent (due to the pre-trained language model) and they refer to visual elements (due to the image-text matching). 
Moreover, the framework can argue with common sense, i.e.\ it states that \emph{a train station is the only place the train goes}.
It combines this prior knowledge from the pre-trained language model with visual concepts detected by the VQA model, such as identifying that the masked image shows \emph{a school bus}.
The capability to mention visual elements stems from the attention controlled guiding of the language generation.
Additionally, the translations accurately describe the part of the image that the VQA model used to answer the question (e.g.\ the pickup truck in the top left example).

An observed failure pattern can be seen in the bottom-right example in \Cref{fig:qualitative-examples}.
It seems very plausible that the person would be \emph{tired}. However, this sentence does not refer to any visual information but instead uses common sense to explain the answer to the input question.
In conclusion, the qualitative results in \Cref{fig:qualitative-examples} show that \modelName indeed generates text which mentions visual information contained in the attention patterns that are extracted from the VQA model. Additionally, the generated sentences are overall grammatically correct and fluent.

\section{Limitations}
Our proposed framework translates the internal attention of a VQA model into natural language. As no datasets exist specifically for this task, we chose to automatically evaluate our text translations of attention maps on textual explanation datasets.
Due to the inherent task differences, we do not expect our attention translations to perfectly match the ground-truth explanations (e.g.\ in terms of writing style), explaining part of the performance gap compared to supervised upper bounds in \Cref{tab:sota-results}. Further research into zero-shot translation methods and the creation of attention translation datasets will be important to better understand attention-based models using natural language.

Our \emph{translation} approach relies on (pre-)trained language models. As a result, the faithfulness of the generated text with respect to the task model is hard to quantify.
To address this, we use an attention-controlled visual guiding component to align the text generation with the VQA model.

Furthermore, we have only considered the ALBEF VQA model. However, our approach could easily be extended to other models such as LXMERT~\cite{tan2019lxmert} or ViLT~\cite{Kim2021ViLTVT}.

Our attention-controlled guiding outperforms guiding with the full image by only a slight margin (c.f.~\Cref{tab:image-masking-ablation}). This could be due to wrong internal reasoning of the task model and attention rollout not identifying the areas causing this. This should be addressed in future work, e.g.\ by using attention-perturbation to understand the importance of image patches.

Lastly, we hypothesize that the relatively small changes in the ablation studies might be due to the LLM already predicting a common sense translation for the question and answer.

\section{Conclusion}
In this work, we introduce \modelName, a zero-shot framework for translating the aggregated attention in a VQA model to natural language. 
In particular, the language generation is guided using the VQA model itself, by means of its internal attention combined with its image-text matching capabilities for selecting word candidates in the language generation.
Our proposed method does not require any training and can be flexibly used together with any language model to translate the visual attribution output for an attribution method of choice. 
Our framework outperforms zero-shot image captioning baselines on textual explanation datasets in the context of VQA\@.

\vspace{1em}
\mypara{Acknowledgements.}
The authors thank IMPRS-IS for supporting Leonard Salewski. This work was partially funded by the Max Planck Society, the BMBF Tübingen AI Center (FKZ: 01IS18039A), DFG (EXC number 2064/1 – Project number 390727645), ERC (853489-DEXIM), and DFG-CRC 1233 (Project number 276693517).

%
%
%
\bibliographystyle{splncs04}
\bibliography{bib}

\newpage
\setcounter{section}{0}
\renewcommand*{\theHchapter}{chX.\the\value{chapter}}
\renewcommand\thesection{\Alph{section}}
\newcommand{\placeholder}[1]{\textlangle{}#1\textrangle{}}
\begin{center}
\large{\textbf{Supplementary Material:\\ Zero-shot Translation of Attention Patterns in VQA Models to Natural Language}}
\end{center}

\ifreview
\else

    \authorrunning{L.\ Salewski et al.}
\fi


In this supplementary material, we show the full prompt templates used for our proposed \modelName in \Cref{sec:full-prompt-templates} and additional translation quality results with different evaluation protocols in \Cref{sec:protocols}.
Additionally, we compare the inference speed of all methods in \Cref{sec:inference-speed}, ablate guiding related parameters in \Cref{sec:guiding-parameters} and show the setup for the in-context example prompting in \Cref{sec:few-shot}.

\section{Full Prompt Templates}\label{sec:full-prompt-templates}
In this section, we show the full prompt templates abbreviated in Section 4.5 and Table 5 of our main paper. In the abbreviations, we used \textlangle{}q\textrangle{} and \textlangle{}a\textrangle{} as placeholders for the corresponding question and answer respectively.

\mypara{``\textlangle{}q\textrangle{}? The answer is \textlangle{}a\textrangle{} because''.} 
One example prompt for this variant would be ``Is that healthy? The answer is no because''.
Here the question \textlangle{}q\textrangle{} is ``Is that healthy'' and the answer \textlangle{}a\textrangle{} is ``no''.

\mypara{\enquote{Q: \textlangle{}q\textrangle{}? A: \textlangle{}a\textrangle{}. E:}.}
The unabbreviated version of this prompt is \enquote{Question: \textlangle{}q\textrangle{}? Answer: \textlangle{}a\textrangle{}. Explanation:}.
An example of this prompt would be \enquote{Question: Is that healthy? Answer: no. Explanation:},
where the question \textlangle{}q\textrangle{} is ``Is that healthy''. and the answer \textlangle{}a\textrangle{} is ``no''.

\mypara{\enquote{Q: \textlangle{}q\textrangle{}?\texttt{\textbackslash n} A: \textlangle{}x\textrangle{}.\texttt{\textbackslash n} E:}.}
The complete version of this prompt variation is \enquote{Question: \textlangle{}q\textrangle{}\texttt{\textbackslash n}? Answer: \textlangle{}x\textrangle{}\texttt{\textbackslash n}. Explanation:}.
In this prompt, \texttt{\textbackslash n} is used to denote a new line.
If the question \textlangle{}q\textrangle{} is ``Is that healthy'' and the answer \textlangle{}a\textrangle{} is ``no'', one example prompt would be \enquote{Question: Is that healthy?\texttt{\textbackslash n} Answer: no.\texttt{\textbackslash n} Explanation: }.

Additionally, we test three prompts with an explicit task instruction that precedes the question and answer.

\mypara{\enquote{Explain the A: \textlangle{}q\textrangle{}? The A is \textlangle{}a\textrangle{} because}}
is the abbreviated version of this prompt \enquote{Explain the Answer: \textlangle{}q\textrangle{}? The answer is \textlangle{}a\textrangle{} because}.  
One example of this prompt would be \enquote{Explain the Answer: Is that healthy? The answer is no because}.

\mypara{\enquote{A and Explain: \textlangle{}q\textrangle{}?\texttt{\textbackslash n} The A is \textlangle{}a\textrangle{} because}.}
The full version of this variant is \enquote{Answer and Explain: \textlangle{}q\textrangle{}?\texttt{\textbackslash n} The answer is \textlangle{}a\textrangle{} because}.
The \texttt{\textbackslash n} character denotes a new line.
For the question \textlangle{}q\textrangle{} ``Is that healthy'' and the answer \textlangle{}a\textrangle{} ``no'', this prompt would be \enquote{Answer and Explain: Is that healthy?\texttt{\textbackslash n} The answer is no because}.

\mypara{\enquote{A and Explain: \textlangle{}q\textrangle{}? The A is \textlangle{}a\textrangle{} because}.}
The expanded version of this variant is \enquote{Answer and Explain: \textlangle{}q\textrangle{}? The answer is \textlangle{}a\textrangle{} because}.
An instance of this prompt would be \enquote{Answer and Explain: Is that healthy? The answer is no because}.

As the last prompt performed best in our VQA-X ablations we used it for all our experiments unless noted otherwise.

\section{Different Evaluation Protocols}\label{sec:protocols}

{
\renewcommand{\arraystretch}{1.2}
\begin{table*}[t]
    \centering
    \resizebox{\linewidth}{!}{
        \begin{tabular}{ l l H H H c c c c c  c  H H H c c c c c}
            \toprule
             & & \multicolumn{8}{c}{\textbf{GQA-REX}~\cite{hudson2019GQANewDataset,Chen2022REXRA}}  & & \multicolumn{8}{c}{\textbf{VQA-X}~\cite{hukpark2018MultimodalExplanationsJustifying}}  \\
            \cmidrule(lr){3-10} \cmidrule(lr){12-19}
            \textbf{Framework}$\downarrow$ & Evaluation &  B1 & B2 & B3 & B4 & M & RL & C & S  & &  B1 & B2 & B3 & B4 & M & RL & C & S \\
            \midrule
            ZeroCap \textsubscript{GPT2}~\cite{Tewel2021ZeroCapZI}$^*$ & GT conditioned & 10.6 & 4.7 & 2.4 & 1.4 & 4.6 & 12.3 & 16.9 & 5.3 && 10.3 & 4.3 & 1.7 & 0.7 & 4.7 & 14.0 & 5.8 & 2.0  \\
            EPT \textsubscript{GPT2}~\cite{Tewel2022ZeroShotVC}$^*$ & GT conditioned & 4.5 & 1.0 & 0.3 & 0.0 & 3.3 & 3.2 & 2.6 & 2.8 && 20.8 & 6.4 & 2.3 & 0.9 & 6.5 & 14.9 & 6.7 & 2.9 \\
            MAGIC \textsubscript{GPT2}~\cite{Su2022LanguageMC}$^{*}$ & GT conditioned & 18.5 & 7.9 & 4.0 & 2.3 & 10.8 & 18.8 & 41.1 & 18.8 && 26.1 & 9.0 & 3.1 & 1.0 & 8.8 & 19.3 & 10.6 & 7.1  \\
            MAGIC \textsubscript{OPT 6.7B}~\cite{Su2022LanguageMC}$^{*}$ & GT conditioned & 23.0 & 10.8 & 5.6 & 3.3 & 11.6 & 22.2 & 48.8 & 21.4 && 27.5 & 10.6 & 4.3 & 1.9 & 9.5 & 20.5 & 14.7 & 8.9\\
            Socratic Models \textsubscript{OPT 6.7B}~\cite{Zeng2022SocraticMC}$^{*}$ & GT conditioned & 20.3 & 9.7 & 5.4 & 3.3 & 14.1 & 22.8 & 40.5 & 19.3 && 31.3 & 15.4 & 7.6 & 3.6 & 12.8 & 25.7 & 19.9 & 10.1\\
            \midrule
            \modelName \textsubscript{OPT 6.7B} (ours) & All  & 38.5 & 22.6 & 14.7 & 10.2 & 18.2 & 35.0 & 113.5 & 31.4 && 41.6 & 25.5 & 14.7 & \bf 8.5 & \bf 13.8 & \bf 34.2 & \bf 38.1 & \bf 10.5\\
            \modelName \textsubscript{OPT 6.7B} (ours) & GT conditioned & 40.3 & 23.7 & 15.7 & 11.0 & 19.4 & 36.6 & 122.8 & 34.4 && 41.5 & 25.2 & 14.4 & 8.3 & 13.7 & 34.0 & 37.4 & 10.4 \\
            \modelName \textsubscript{OPT 6.7B} (ours) & Answer correct & \bf 41.5 & \bf 24.9 & \bf 16.8 & \bf 12.0 & \bf 20.6 & \bf 38.6 & \bf 136.5 & \bf 36.7 && \bf 41.6 & \bf 25.5 & \bf 14.8 & \bf 8.5 & \bf 13.8 & \bf 34.2 & 37.9 & 10.4\\
            \bottomrule
        \end{tabular}
    }
    \caption{
    Zero-shot attention rollout~\cite{abnar2020quantifying} to natural language translation results in the context of VQA on GQA-REX and VQA-X on the subset of correctly answered questions. We report Bleu-1~(B1), Bleu-2~(B2), Bleu-3~(B3), Bleu-4~(B4), Meteor~(M), Rouge-L~(RL), Cider~(C), and Spice~(S). Higher is better for all reported metrics. \emph{GT conditioned} evaluates on the entire test split but conditions the language model on the ground-truth answer. \emph{Answer correct} only evaluates on the subset of translations where the VQA model predicted the answer correctly. $^*$We adapted the related frameworks to the task, and also provided them with privileged access to the ground-truth answers (thus their scores are unchanged).
    }%
    \label{tab:sota-results-answer-correct}
\end{table*}
}

Here, we provide additional quantitative results for generated translations of attention patters on the VQA-X and GQA-REX datasets.
When evaluating texts that are generated based on predicted answers, several different evaluation schemes are possible~\cite{hukpark2018MultimodalExplanationsJustifying,wu2019FaithfulMultimodalExplanation,Sammani2022NLXGPTAM,kayser2021vil}.

First, the quality metrics can only be reported on the subset of texts for which the answer was correctly predicted (sometimes also referred to as \emph{filtered}).
An advantage of this evaluation is that it is not clear what the matching translation of an incorrect answer should contain and how to evaluate it, e.g.\ should the translation be wrong when the answer is wrong, what should be wrong about it?
A disadvantage of this evaluation is that the number of samples and the samples that are used in the evaluation vary for different methods.

Second, the models can be conditioned on the ground-truth answer. 
This makes the number of samples comparable amongst different methods, however some internal states of the models may still be incorrect.

The third option is to report the metrics for all samples independent of the correctness of the answer. 
We chose this option in the main paper and only discuss the results for the results for the additionally evaluation schemes here.

Note, that the metrics of the related zero-shot works~\cite{Tewel2021ZeroCapZI,Dathathri2019PlugAP,Tewel2022ZeroShotVC,Zeng2022SocraticMC} as reported in the main paper do not change, as they were already conditioned with the ground-truth answer and thus \enquote{answered} all samples correctly.

On VQA-X~\cite{hukpark2018MultimodalExplanationsJustifying}, our base VQA model ALBEF~\cite{Li2021AlignBF} achieves an answer accuracy of 96.2\%.
Hence, the subset of correctly answered questions is nearly identical to the full test set (as only 74 answer predictions are wrong).
As a result, restricting the evaluation to the correctly answered samples only does not affect the scores much and there are only minor changes in the quality of the translated attention patterns.
All four Bleu metrics only differ by up to $0.1$, the Meteor, Rouge-L and Spice values remain unchanged.
The Cider value changes by $0.2$.
Analogously, the results for the ground-truth conditioned variant are also very similar.
All values are listed in \Cref{tab:sota-results-answer-correct}.

On GQA-REX~\cite{hudson2019GQANewDataset,Chen2022REXRA}, our base VQA model ALBEF~\cite{Li2021AlignBF} achieves an answering accuracy of 57.9\%.
On the subset of correctly answered questions, the scores of our \modelName framework are slightly higher across all metrics.
For example Bleu-4 goes from 10.2 to 12.0, Meteor goes from 18.2 to 20.6.
The largest increase is in Cider which increases from 113.5 to 136.5.
Similarly, all other metrics are larger if evaluated on the subset of correctly answered questions.
As a consequence of this slight increase, within the only correctly answered evaluation protocol, the gap in performance between our \modelName framework and the related works is even bigger.
The ground-truth conditioned values (\emph{GT conditioned}) are higher than the \emph{All} values, which indicates that the correct answer leads to a better translation of the salient image regions.
However, they are well below the \emph{answer correct} scores, which shows that not only the correct answer prediction but also the selection of the correct image regions that are responsible for the correct answer prediction are important for the translation.
All metric values are listed in \Cref{tab:sota-results-answer-correct}.

\section{Inference Speed}\label{sec:inference-speed}
\begin{wraptable}[13]{r}{.5\linewidth}
\vspace{-2em}
    \centering
    \resizebox{\linewidth}{!}{
        \begin{tabular}{l c}
            \toprule
            Framework & Seconds / Iteration \\
            \midrule
            ZeroCap \textsubscript{GPT2}~\cite{Tewel2021ZeroCapZI}$^*$ & 32.1 \\
            EPT \textsubscript{GPT2}~\cite{Tewel2022ZeroShotVC}$^*$ & 91.2 \\
            MAGIC \textsubscript{OPT 6.7B}~\cite{Su2022LanguageMC}$^{*}$ & \bf 1.2\\
            Socratic Models \textsubscript{OPT 6.7B}~\cite{Zeng2022SocraticMC}$^{*}$ & 6.0\\
            \midrule
            \modelName \textsubscript{OPT 6.7B} (ours) & 7.2\\
            \bottomrule
        \end{tabular}
    }
    \caption{
    Average inference speeds for all compared methods. Lower is better. $^*$We adapted the related frameworks to the task.
    }%
    \label{tab:inference-speed}
\end{wraptable}

As mentioned in the main paper, the inference speed of the different methods vary greatly.
For completeness, we list the inference speeds for all methods presented in the main paper in \Cref{tab:inference-speed}.
For this compute the average time needed per sample on a NVIDIA V100 GPU\@.

ZeroCAP~\cite{Tewel2021ZeroCapZI} and EPT~\cite{Tewel2022ZeroShotVC} are the slowest methods with 32.1 and 91.2 seconds per iteration respectively.
This is the case because they use backpropagation and optimization of the hidden states of the large language model to control the text generation.

On the other hand, MAGIC~\cite{Su2022LanguageMC} is the fastest method with only 1.2 seconds per iteration.
Socratic Models~\cite{Zeng2022SocraticMC} is only a bit slower with 6.0 seconds per iteration, which can be attributed to it generating multiple candidate sentences, which are then all ranked by CLIP\@.

Our method \modelName needs 7.2 seconds per iteration, but in contrast to the above method this also includes the inference of the VQA model.
We find that our method significantly outperforms the next best approach Socratic Models in terms of natural language generation metrics (c.f. \Cref{tab:sota-results-answer-correct}) at only a marginal slow down in terms of inference speed.

\section{Ablating Guiding Parameters}\label{sec:guiding-parameters}
In this section, we ablate the guiding parameters $\tau$ and $\kappa$, which influence how the VQA model guides the language model.

\begin{figure}
     \centering
     \begin{subfigure}[t]{0.475\textwidth}
    \centering
    \includegraphics[width=\linewidth]{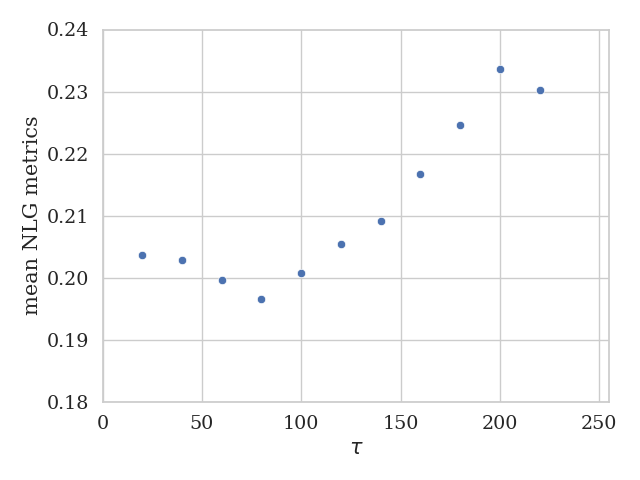}
     \end{subfigure}
     \hfill
     \begin{subfigure}[t]{0.475\textwidth}
    \centering
    \includegraphics[width=\linewidth]{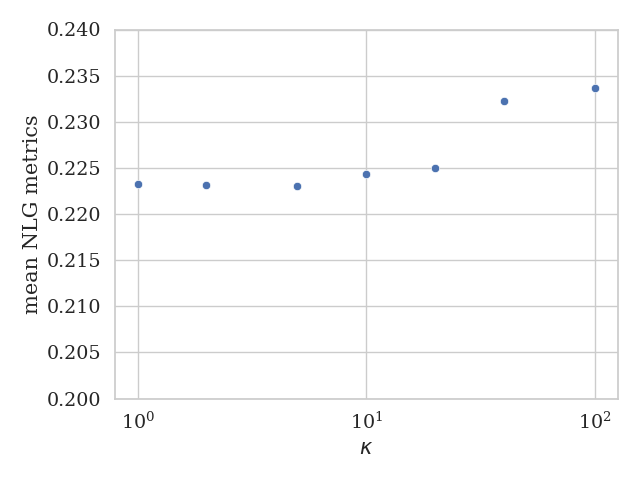}
     \end{subfigure}
    \caption{Ablating the effect of different attention rollout thresholding values $\tau$ (left) and of different values for $\kappa$ (right) on VQA-X. The y-axis is the mean of all NLG metrics (B1-B4, M, RL, C and S).}%
    \label{fig:tau-kappa-ablations}
    \vspace{-0.4ex}
\end{figure}

First, we investigate the effect of the thresholding parameter $\tau$.
It is used in Equation 6 of the main paper to compute the binary mask which is applied to the image. 
We show the effects of varying $\tau$ in \Cref{fig:tau-kappa-ablations} (left) and find that a value of $\tau=200/255$ works best with the average of the NLG metrics being 23.4 on VQA-X.

On the other hand, the hyperparameter $\kappa$ controls the pre-softmax temperature in Equation 8 of the main paper.
The larger the value, the more emphasis is put on a singular visual concept candidate proposed by the language model.
In our main paper we used $\kappa=100$.
In \Cref{fig:tau-kappa-ablations} (right), we show the results for different values of $\kappa$ on VQA-X.
It can clearly be seen, that as hypothesized, larger values between $\kappa=40$ and $\kappa=100$ are needed to obtain better performance.
Larger values sharpen the output of the softmax and thus give the VQA model, used to quantify image-text matching, more influence in selecting specific tokens during the guiding process.

\section{Impact of In-Context Generation ($n$-Shot Prompting)}\label{sec:few-shot}
Here we give more details for the in-context generation.
We purely rely on in-context demonstration of examples and do not use an instructional prompt for the language model.
First we repeat $n$ random examples from the training set and use the following prompt (without the line break).
Expressions like \placeholder{question} are placeholders that are replaced at run time with their respective values from the dataset samples.

\footnotesize{
\begin{align*}
    \underbrace{\text{Question: \placeholder{question}? Answer: \placeholder{answer}. Explanation: \placeholder{explanation}.}}_{n\text{ in-context examples}} \\
    \text{Question: \placeholder{question}? Answer: \placeholder{predicted answer}. Explanation:}
\end{align*}
}

\end{document}